\documentclass[letterpaper,journal]{IEEEtran}

\usepackage{amsmath,amsfonts}
\usepackage{amssymb}
\usepackage{algorithmic}
\usepackage{algorithm}
\usepackage{balance}
\usepackage{array}
\usepackage[caption=false,font=normalsize,labelfont=sf,textfont=sf]{subfig}
\usepackage{textcomp}
\usepackage{stfloats}
\usepackage{url}
\usepackage{verbatim}
\usepackage{graphicx}
\usepackage{cite}
\usepackage{acronym}
\usepackage{orcidlink}
\usepackage{colortbl}
\usepackage{threeparttable}
\usepackage{parskip}
\usepackage{xcolor}
\usepackage{enumitem}
\usepackage{hyperref}
\usepackage{siunitx}
\usepackage{booktabs}
\usepackage[caption=false,font=normalsize,labelfont=sf,textfont=sf]{subfig}
\usepackage{amsthm}

\colorlet{RED}{red}
\colorlet{BLACK}{black}

\newcommand{\figref}[1]{Fig. \ref{#1}}
\newcommand{\secref}[1]{Section \ref{#1}}
\newcommand{\tabref}[1]{Table \ref{#1}}
\newcommand{\ie}[0]{i.e.~}
\renewcommand{\quote}[1]{``#1''}

\newif\ifappendix
\appendixtrue   

\definecolor{layer400_800}{HTML}{FF8C00}
\definecolor{layer800_512}{HTML}{00FF00}
\definecolor{layer512_256}{HTML}{FF00FF}
\definecolor{layer256_10}{HTML}{FFFF00}
\definecolor{min_network}{HTML}{00BFFF}

\begin{document}
\acrodef{ai}[AI]{Artificial Intelligence}
\acrodef{ml}[ML]{Machine Learning}
\acrodef{snn}[SNN]{Spiking Neural Network}
\acrodef{cnn}[CNN]{Convolutional Neural Network}
\acrodef{ann}[ANN]{Artificial Neural Network}
\acrodef{bnn}[BNN]{Binarized Neural Network}
\acrodef{shd}[SHD]{Spiking Heidelberg Digits}
\acrodef{ssc}[SSC]{Spiking Speech Commands}
\acrodef{psmnist}[PSMNIST]{Permuted Sequential-MNIST}
\acrodef{sota}[SOTA]{state-of-the-art}
\acrodef{lif}[LIF]{Leaky-Integrate and Fire}
\acrodef{lod}[LOD]{Leading-One Detector}
\acrodef{lopd}[LOPD]{Leading-One Position Detector}
\acrodef{noc}[NoC]{Network-on-Chip}
\acrodef{fpga}[FPGA]{Field-programmable Gate Array}
\acrodef{asic}[ASIC]{Application-specific Integrated Circuit}
\acrodef{aer}[AER]{Address Event Representation}
\acrodef{orf}[ORF]{OpenROAD-Flow-Script}
\acrodef{pdk}[PDK]{Process Design Kit}
\acrodef{pe}[PE]{Processing Element}
\acrodef{dte}[DTR]{Differential Time Representation}
\acrodef{dtr}[DTR]{Differential Time Representation}
\acrodef{ips}[IPS]{Inferences per second}
\acrodef{alm}[ALM]{Adaptive Logic Module}
\acrodef{clb}[CLB]{Configurable Logic Block}
\acrodef{lut}[LUT]{Look-Up Table}
\acrodef{dsp}[DSP]{Digital-Signal Processing}

\newtheorem{proposition}{Proposition}
\newtheorem{theorem}{Theorem}[section]

\title{Low-Power Spiking Neural Network Acceleration with Differential Time Representation: A Modular Hardware Framework}

\author{Daniel Windhager$^{\ast\P}$\orcidlink{0009-0007-6443-6515}, Lothar Ratschbacher$^\dagger$\orcidlink{0000-0002-2631-0977}, Bernhard A. Moser$^{\ast\ast\ddagger}$\orcidlink{0000-0003-1859-046X}, Michael Lunglmayr$^\S$\orcidlink{0000-0002-4014-9681}
\thanks{$\hspace{-1em}^\ast$Intelligent Wireless Systems, Silicon Austria Labs, Linz, Austria, daniel.windhager@silicon-austria.com\\
$^{\P}$double affiliation: Institute of Signal Processing, Johannes Kepler University, Linz, Austria\\
$^\dagger$Intelligent Wireless Systems, Silicon Austria Labs, Linz, Austria, lothar.ratschbacher@silicon-austria.com\\
$^\ddagger$Institute of Signal Processing, Johannes Kepler University, JKU SAL IWS Lab, Linz, Austria, bernhard.moser@jku.at\\
$^{\ast\ast}$double affiliation: Software Competence Center Hagenberg (SCCH), 4232 Hagenberg, Austria\\
$^\S$Institute of Signal Processing, Johannes Kepler University, Linz, Austria, michael.lunglmayr@jku.at}
}

\markboth{IEEE TRANSACTIONS ON VERY LARGE SCALE INTEGRATION (TVLSI)}%
{Windhager \MakeLowercase{\textit{et al.}}: \title}

\maketitle

\begin{abstract}
Spiking Neural Networks (SNNs) offer promising energy efficiency, making them well suited for energy-constrained applications. Much of their energy savings stems from the sparse encoding of spike trains. We show that spike trains can be represented more efficiently using Differential Time Representation (DTR), which requires fewer bits than Address-Event Representation (AER) or equidistantly sampled 1-bit spike trains. To support multiple input spike trains, we provide a rigorous proof that DTR sequences can be merged with low complexity. We also provide a set of modular hardware streaming components using this representation that can be flexibly connected to construct custom accelerator architectures for specific SNNs. Tailoring the hardware architecture to the selected SNN is streamlined by a Python framework combining the modular hardware components into a full SNN implementation. We use the framework to generate three accelerators for the Spiking Heidelberg Digits (SHD), Spiking Speech Commands (SSC), and PSMNIST benchmarks, each in a throughput-optimized parallel variant and a resource-reduced time-multiplexed variant. We report results for FPGA and ASIC implementations, the latter synthesized with a fully open-source flow targeting the predictive ASAP7 PDK. On the SHD dataset, the ASIC implementation reaches $92.46\%$ classification accuracy, while sustaining $\approx{}12{,}027$ inferences per second at $20.37\mu{}$J per inference, exceeding the neuromorphic Loihi processor by $109\times$ in throughput and $1.8\times$ in energy efficiency. The corresponding FPGA implementation achieves $\approx{}6{,}023$ inferences per second without using DSP blocks. Across all three datasets, the synthesized ASICs reduce energy per inference by $27$--$37\times$ relative to their FPGA counterparts, demonstrating that competitive SNN accelerators can be realized entirely within an open-source design flow.
\end{abstract}

\begin{IEEEkeywords}
Spiking Neural Networks, Differential Time Representation, Modular hardware, Streaming, FPGA, ASIC
\end{IEEEkeywords}

\section{Introduction}
\IEEEPARstart{T}{he} field of \ac{ai} and more specifically \ac{ml} is evolving, with a clear trend towards larger and more expansive network architectures, consequently leading to larger datasets and parameter spaces. While GPUs, faster memory and higher bandwidths can alleviate the impact of high memory requirements for server-side applications, large models present a significant challenge in embedded and edge applications where power efficiency and computational constraints are paramount. Accordingly, less computationally expensive architectures have gained interest in academia and industry in recent years. One of the approaches is to utilize \acp{snn} instead of traditional \acp{ann}, due to their improved power efficiency~\cite{yan2026}, sparsity characteristics~\cite{Moser_2025} and, depending on the choice of the neuron model and spike encoding, decreased computational effort. This is especially important for embedded devices and edge applications, as these often impose hard limits on energy efficiency and inference time. 

Notably, a significant part of the efficiency advantage that \acp{snn} have over \acp{ann} is due to the sparse nature of their inputs, as well as efficient data representation schemes used for these spike trains\cite{snn_vs_ann}. This representation is particularly important, since it not only influences the number of symbols used to encode a spike train, but also directly shapes the computation required to process these spikes. Therefore, we advocate for the use of \acf{dte}~\cite{aisys} over other commonly used schemes such as the absolute time representation often employed by the \ac{aer}, as well as the equidistantly sampled \mbox{1-bit} spike trains. We demonstrate that \ac{dte} outperforms both aforementioned representations in terms of total number of bits required. Furthermore, we also highlight the fact that the differential time used for \ac{dte} can be directly leveraged for computing neuron dynamics. 

Unlike absolute time representation, \ac{dte} requires consideration of spike histories when comparing spikes across different spike trains. We present an efficient algorithm that enables the merging of \ac{dte} spikes across different spike trains, without the need to convert to other representations. Furthermore, we demonstrate the implementation of this algorithm in digital hardware, showing its practical feasibility.

Based on \ac{dte}, we present a suite of resource-efficient, extensible and configurable hardware components. These can be connected flexibly, enabling the implementation of hardware accelerator architectures tailored to state-of-the-art \acp{snn}. The hardware components include fully connected blocks, recurrent path blocks and delay blocks, as well as an implementation for efficiently merging spikes across different spike trains.

 With these components, arbitrary feedforward and recurrent spiking neural network topologies, with axonal delays in the recurrent or in the forward path can be constructed. Furthermore, while our implementations use the standard \ac{lif} neuron model for activation after the linear layer, other neuron models can be used instead (through the substitution of a single VHDL entity). Another advantage of this modular approach for each layer in the network is the ability to quantize each layer individually, minimizing quantization-induced noise even when employing lower bit width. This enables a wide array of possible network architectures, e.g. the network that achieves state-of-the-art accuracy on the \ac{ssc} dataset\cite{shd_dataset}. 

To facilitate the rapid configuration of multiple networks, we provide a Python framework that generates the corresponding top-level VHDL files necessary to instantiate all components. The networks can then be defined using a JSON file specifying the individual layers in the \ac{snn}, as well as their respective characteristics, such as quantization bit widths.

We performed extensive synthesis for both FPGA and ASIC implementations. The performance metrics of the synthesized designs are benchmarked against current state-of-the-art hardware accelerators, demonstrating the feasibility of the designs and the overall approach.

To summarize, the main contributions of this work are: 
\begin{itemize}
    \item \textbf{Theoretical and Empirical Justification for \ac{dte}:} Demonstrates \ac{dte}'s superiority over \ac{aer} and sampled spikes for modular streaming architectures through comparative analysis (\secref{sec:dte}).
    \item \textbf{Mathematical Foundation:} Provides rigorous proofs enabling direct comparison of differential-time domain spike trains (\secref{sec:dte}\ifappendix, Appendix\fi).
    \item \textbf{Modular Hardware Suite:} Configurable VHDL components supporting arbitrary \ac{snn} topologies (including recurrence and axonal delays), per-layer quantization, and efficient local SRAM utilization (\secref{sec:hw_arch}).
    \item \textbf{Automated Design Flow:} A Python tool auto-generates top-level VHDL from JSON configurations, streamlining \ac{snn}-to-hardware implementation (\secref{sec:reprod}).
    \item \textbf{Benchmarking and Validation:} Evaluates performance and accuracy on SHD, SSC, and PSMNIST datasets, demonstrating state-of-the-art results across diverse spiking benchmarks (\secref{sec:results}).
    \item \textbf{Synthesis Results:} Reports area, power, and performance metrics for the ZCU102 FPGA and preliminary 7nm ASAP7 ASIC implementations using open-source tools (\secref{sec:results}).
\end{itemize}

\section{Background and related work}
\label{sec:bg_rl_work}

In recent years, researchers and engineers have adopted \mbox{FPGAs} for accelerating neural networks in cases where GPUs cannot be used due to power or form factor constraints. A significant number of publications have proposed various FPGA-based accelerator architectures for \acp{snn}, either for specific applications or designed as general-purpose accelerators\cite{deepfire2,liu_hanwen,firefly,syncnn,s2n2,nevarez}. However, many of these designs feature similarities to existing CPU or GPU architectures. They consist of a central controller coordinating the acceleration process, aided by many smaller distributed \acp{pe}, which are responsible for carrying out the computation \cite{sibrain,comp_eff_acc_li,spiking_transformer_3d,noc_based_snn,nimbekar_hw_sw_co}. These \acp{pe} are often arranged in lattice-like structures, similar to the systolic arrays found in modern GPUs, making them highly effective for dense matrix operations, but less so for the inherent temporal sparsity found in \acp{snn}. 

Due to this flexibility in the design, however, the need for routing data and inputs (\ie spikes) to their respective \acp{pe} becomes a critical challenge. Notably, much of the power for modern accelerators is consumed by data transfers, instead of computation\cite{horowitz_computing,8993434}. Therefore, in order to enable the next generation of edge devices to use more capable models, it is necessary to limit the volume of data transfers involved. To optimize efficiency, custom accelerators can provide relief by carefully utilizing local memories. 

Another important consideration when minimizing data transfers and subsequently also processing power (due to transfers as well as computations), is the choice of representation of the inputs to the accelerator, \ie the spikes and spike trains entering the system. Mathematically, spikes are Dirac delta pulses, delayed by a specific time (\ie a number of cycles). A sequence of these spikes constitutes a spike train. Various different representations of these spike trains exist in hardware, each with its respective benefits and drawbacks. Note that this is entirely separate from coding methods such as Rate Coding or Latency Coding\cite{snntorch}, which are solely used for transforming analog data or natural vectors into spike trains. Spike representation focuses on how to represent a given spike train efficiently, \ie minimizing the bit-width overhead for the target hardware. Existing literature frequently uses the \ac{aer}\cite{liu_pan,mesquida,kim_park,madhuvanthi}, which can be inefficient as it often uses absolute time to store the delays of each of the spikes. The \ac{dte} method presented in \cite{aisys} is a more efficient method to represent spikes, as will be demonstrated in the next section.

\section{Efficient spike representation for digital hardware}
\label{sec:dte}
\begin{figure}
    \centering
    \includegraphics[width=\linewidth]{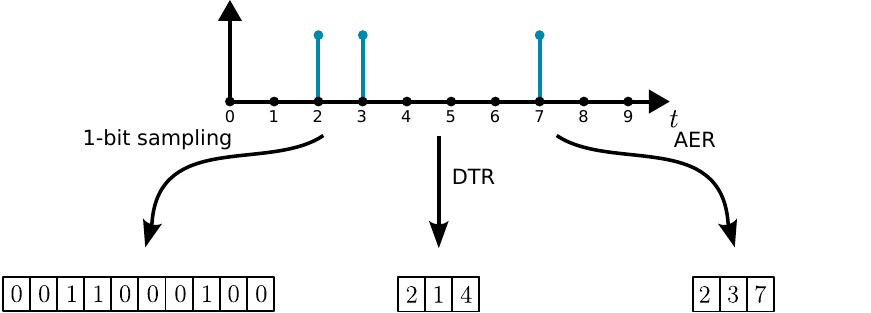}
    \caption{Comparison of a sample spike train represented using 1-bit direct sampling, \acf{dte} and \acf{aer}. For 1-bit sampling the entire spike train is sampled equidistantly producing a series of 1-bit values representing the spike train. \ac{aer} stores the time of occurrence (from some predefined starting point) of each spike. \ac{dte} only stores the time difference between spikes, potentially requiring fewer bits than \ac{aer} and fewer values than the 1-bit sampling, as shown by this example where 1-bit sampling requires 10 bits, \ac{dte} requires 9 bits and \ac{aer} requires 12 bits.}
    \label{fig:dte_aer_sampling}
\end{figure}
This section evaluates \acf{dte} as a resource-efficient alternative for representing individual spikes and spike trains in digital hardware. Note that this step of differential time representation occurs after the initial sampling of the data and after the coding of the data (e.g. using rate coding or latency coding), and is thus largely independent of both. While one could argue that individually representing spikes is wasteful due to the fact that many works use rate coding and therefore entire spike trains can be represented with a single scalar denoting the firing rate, there are works that showcase the benefit of using other methods that rely on the temporal characteristics of individual spikes to encode information\cite{11460632,9747906}.

We benchmark \ac{dte} against the two most widely adopted methods in current literature: \acf{aer} and equidistant \mbox{1-bit} sampling. In traditional \ac{aer}, a spike is typically represented by its source or destination address, often augmented by an absolute timestamp and polarity information. While robust, this can lead to significant data overhead in dense spike trains. Conversely, 1-bit sampling represents the spike train as a continuous stream of bits where each bit corresponds to a fixed time step ('1' denoting the presence of a spike, '0' the absence of a spike). While simple to implement, this method scales poorly with high temporal resolution. As illustrated in \figref{fig:dte_aer_sampling}, \ac{dte} optimizes this process by encoding only the temporal intervals (delta time) between successive spikes. If the bit width is insufficient to encode the time difference, the maximum encodable time difference can be defined as an overflow symbol, denoting that more than the maximum representable time has passed. The accumulation of this maximum delta time with the following delta times, up to the point where a time difference smaller than the maximum is encountered, reconstructs the delta time in such a case. We show that, by focusing on relative rather than absolute timing, \ac{dte} significantly reduces the required bit width compared to \ac{aer} and minimizes the redundant data inherent in \mbox{1-bit} sampling.

\subsection{DTR vs AER}
The \ac{aer}, being suboptimal for several reasons, can lead to synchronization failures for continuously running systems, as over extended periods of operation the counter often used to represent the current time will overflow. As an alternative, we proposed the \ac{dte} in~\cite{aisys} which is asymptotically equivalent in the worst case, in terms of required bit length for the entire spike train, to the absolute time representation. For most other cases it substantially reduces the bit width required to represent the entire spike train. 

Consider a spike train with $K\in\mathbb{N}$ spikes for which the last spike occurs at some time $T\in\mathbb{N}_0$. Using absolute time (as it is often done for \ac{aer}, as e.g. in~\cite{liu2020effectiveaerobjectclassification,8989987}), $\lceil\log_2(T+1)\rceil$ bits are required for every symbol representing a single spike, whereas the required bit width of a symbol for \ac{dte} depends on the distribution of the number of cycles that pass between spikes (which are typically smaller than the maximum time)\footnote{When thinking about spike sequences without a dedicated time limit, e.g. for architectures performing inference on streaming data, the limitations of an absolute time representation concept immediately become obvious.}. In the absence of overflow symbols, the worst case, where all $K$ spikes occur at the latest possible moment, would also require a bit width of $\lceil\log_2(T+1)\rceil$. If overflow symbols are used, then the worst case for \ac{dte} occurs for long periods without spikes which may cause \ac{dte} to require more bits to represent the entire spike train than absolute time if many overflow symbols are inserted. This can be avoided in practice by choosing the bit width based on the expected distribution of spikes, thereby keeping the number of overflow symbols used to a minimum, while significantly reducing the hardware resources required to store a single spike.

The optimal scenario for \ac{dte} occurs when all $K$ spikes are exactly evenly (\ie equidistantly) spaced, reducing the required bit width to $\lceil\log_2(\frac{T+1}{K})\rceil$. Therefore, the exact savings in terms of bit width depend on the distribution of the spikes. However, even in the worst case the differential time representation at most requires the same number of bits as the absolute time representation. The subsequent analysis incorporates overflow symbols, which further improve the efficiency of \ac{dte}.

\subsection{DTR vs sampled 1-bit spike trains}
An alternative method to represent spike trains is to equidistantly sample the spike trains and use the resulting sequences of ones and zeros (assuming only positive spikes of amplitude one are used), which can be represented as 
\begin{equation}
    S = (s_1, s_2, ..., s_{n-1}), s_i \in \{0, 1\}.
\end{equation}
This sampled version of a spike train is similar to the differential time representation, when the bit width is constrained to a single bit. In a corresponding \ac{dte} representation, 1 would be interpreted as the overflow symbol signifying that one sampling time has passed where no spike was triggered, whereas 0 signals that a spike occurred at the current time step. 

Since spike trains are inherently sparse, \ac{dte} will typically require fewer bits to represent the same spike train than \mbox{1-bit} sampling. Therefore, \ac{dte} offers a favorable balance, in terms of hardware requirements, between using absolute time and equidistant \mbox{1-bit} sampling.

\subsection{Choosing an ideal bit width for DTR}
In scenarios where spike density is low, a larger bit width for the differential time representation would yield an advantage over the \mbox{1-bit} sampled time representation, assuming the bit width is chosen based on the expected distribution of difference times between spikes. An insufficient bit width for a given spike train generates a large number of overflow symbols, thereby reducing efficiency. Furthermore, generating frequent overflow symbols is not only suboptimal in terms of bit utilization, but also increases hardware latency by requiring additional processing cycles. Conversely, an excessively large bit width eliminates overflow symbols but incurs unnecessary overhead when encoding dense spike trains. Therefore, the most efficient bit width must be dependent on the distribution of spikes in the spike trains.

Assuming $b$ is the bit width of a single symbol, then every spike whose time difference $d\in\mathbb{N}_0$ (relative to the previous spike) is less than $2^b-1$, can be represented with a single symbol. The strict inequality is due to one configuration of the bits for the symbol being reserved as the designated overflow symbol. Similarly, every spike with difference $d$ for which $2^b-1 \le d < 2^{b+1}-1$ holds, can be represented with two symbols (one being the overflow symbol). Following the same logic for larger intervals, the number of symbols for a given time difference $d$ and bit width $b$ is
\begin{equation}
    N_b(d) = \left\lceil \frac{d+1}{2^b-1} \right\rceil.
\end{equation}
The $+1$ term in the numerator is necessary to preserve the inequality relationship, for the edge case when $d=0$ or $2^b-1$ divides $d$ without a remainder, due to both $d$ and $2^b-1$ being integers. 

\begin{figure}
    \centering
    \includegraphics[width=\linewidth]{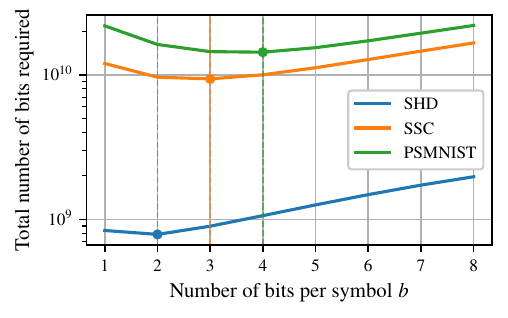}
    \caption{Overall number of bits $B_{\text{total}}$ required to encode all spike trains within the fully trained networks during inference of the whole dataset, respectively, based on chosen bit width $b$ for the differential time. The x-axis shows the number of bits being used to encode the \ac{dte} symbols, while the y-axis shows the overall number of symbols (including overflow symbols) multiplied by the chosen bit width for the three datasets \ac{shd}, \ac{ssc} and \ac{psmnist}. The results are based on the total spike count aggregated across all network layers during inference on the training, validation, and test sets. Vertical lines indicate the optimal bit width $b^*$ that minimizes the total number of bits required for each dataset.}
    \label{fig:ideal_bitwidth}
\end{figure}

Let $\mathcal{D}$ now represent the set of all inter-spike time differences across the entire dataset. Then, the total number of symbols as a function of $b$ is 
\begin{equation}
    N_{\text{total}}(b) = \sum_{d \in \mathcal{D}} N_b(d) = \sum_{d \in \mathcal{D}} \left\lceil \frac{d+1}{2^b-1} \right\rceil.
    \label{eq:global_num_symbols}
\end{equation}
Multiplying this with $b$ leads to the total number of bits required to represent all spikes in the entire dataset, which is expressed as 
\begin{equation}
    B_{\text{total}}(b) = b N_{\text{total}}(b) = b \sum_{d \in \mathcal{D}} \left\lceil \frac{d+1}{2^b-1} \right\rceil.
    \label{eq:global_num_bits}
\end{equation}
Ideally, we want to find the optimal bit width $b^*$ which minimizes \eqref{eq:global_num_bits}, denoted as 
\begin{equation}
    b^* = \underset{b \in \mathbb{N}^+}{\arg\min} \, B_{\text{total}}(b).
    \label{eq:optimal_bit_width}
\end{equation}
Due to the ceiling function, no analytical solution for \eqref{eq:optimal_bit_width} exists. However, a minimum can be determined by empirically evaluating $B_{\text{total}}(b)$ for a finite range of practically relevant $b$-values. The distribution of all differential-time represented spikes that occur in the three datasets for which we evaluate our models (\ac{shd}, \ac{ssc} and \ac{psmnist}) is depicted in \figref{fig:ideal_bitwidth}. To generate this distribution, the train, test and validation splits of each of the datasets were processed by fully trained models while recording all the spikes in the entire network. This also shows that the choice of the ideal bit width depends on the dataset, and consequently, the target application. For example, while a higher bit width $b=4$ minimized the total bit-cost for \ac{psmnist}, indicating that there are long periods in the spike trains where no spikes occur, for \ac{shd} a bit width of $b=2$ is optimal. The ideal bit width being low indicates that most spikes occur in close succession to each other, which further highlights the inefficiency of using absolute time to represent spikes.

\subsection{Comparing spike trains in differential time}
\label{sec:spike_sorter}

\begin{figure}
    \centering
    \includegraphics[width=0.85\linewidth]{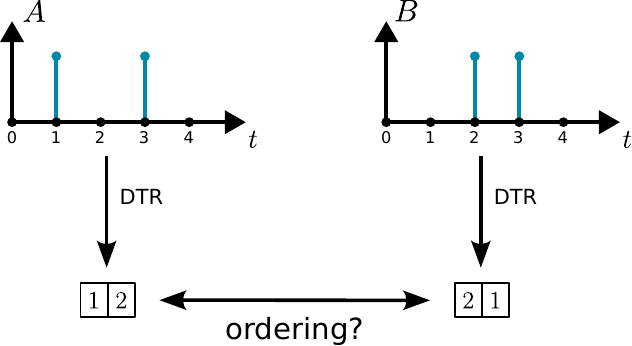}
    \caption{Two separate spike trains given in differential time representation (DTR). From just the representation it is not always immediately obvious which spike (across both spike trains) comes next. While the first spike (from $A$) is readily identified, the next spike is $2$ for both $A$ and $B$, although the actual (absolute) position of the spike for spike train $B$ is at time $t=2$ and thus earlier than the one for spike train $A$ at time $t=3$.}
    \label{fig:spike_sorting}
\end{figure}

When analyzing the three discussed representations \ac{aer}, 1-bit sampled and \ac{dte}, the above results favor \ac{dte}. However, thus far, the analysis has been limited to spikes within a single spike train. In real-world applications, there are many scenarios with multiple separate sources that could produce spike trains, e.g. multiple sensors or ADCs that produce differential time represented spikes. Consequently, a fundamental question arises: how can the next occurring spike be identified when processing multiple \ac{dte} spike trains?

Determining a temporal ordering across multiple spike trains is more challenging for differential time represented spikes than it is for the \ac{aer}. To illustrate this, consider two AER-encoded spikes $a\in\mathbb{N}_0$ and $b\in\mathbb{N}_0$ from two separate spike trains. Without any other information we can determine which spike occurs first, namely $a$ if $a<b$ holds, or $b$ otherwise. 

For the same two spikes represented using \ac{dte}, however, direct comparison is not possible, as their global temporal reference depends on their respective histories, as depicted in \figref{fig:spike_sorting}. Therefore, for \ac{dte}, order is preserved within a spike sequence, but is not guaranteed across different sequences. As we show in the next section, such order can be restored by a hardware-efficient algorithm without the need for conversion to other representations. Furthermore, this algorithm can be computed online, \ie without the need for temporal buffering of incoming spike trains.

The following section demonstrates that multiple differential time represented spike sequences can be merged without explicitly calculating the cumulative sums and subsequently merging the sequences in absolute time. We describe the proof for two sequences, since it can, by induction, be extended to any number of sequences.

Let $ (A =(a_i)_{i=1}^{n}, \leq)$ be the ordered sequence of the absolute spike times of a spike train (the most commonly used ordering of ordered sequences is the natural ordering of time values), and $(B=(b_j)_{j=1}^{m}, \leq)$ an ordered sequence of absolute spike times of a second spike train, defined in the same way. For clarity, all spikes and spike trains given in differential time in this section will be prefixed by $\Delta{}$ to avoid confusion. All other spikes are assumed to be represented in absolute time unless stated otherwise.

The ordered sequences of differential times are then defined such that $(\Delta A =\left(\Delta a_i = a_i-a_{i-1}\right)_{i=1}^{n}, a_{i-1} \leq a_i)$ using $a_0 = 0$.

The merged sequence $Y = {A \cup B}$ is realized by the order relation. Its ordered sequence of differential times $\Delta Y$ is again defined in the same way as described for $\Delta{}A$ above. The cumulative sum of the elements of $\Delta Y$ will again give $Y$. 
\bigskip
\begin{proposition} 
    The ordered sequence $\Delta Y$ can be generated directly from $\Delta A$ and $\Delta B$ using the following procedure: Compare the first elements of the input sequences $\Delta A$ and $\Delta B$ to each other. Append the minimum to the output sequence $\Delta Y$, remove the minimum from its respective sequence, and update the other first element (when using more than two sequences: update all the other first elements) by subtracting the minimum from it. Repeat until all input sequences are empty.
    \label{prop:merge_spikes}
\end{proposition}

\begin{figure}
	\centering
	\includegraphics[width=0.6\linewidth]{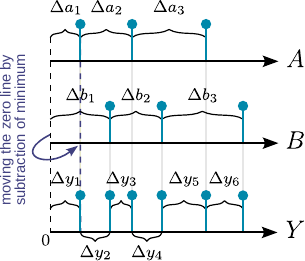}
	\caption{Merging of spike trains given in differential time. \ac{dte} sequence $\Delta{}A=(\Delta{}a_1, \Delta{}a_2, \Delta{}a_3)$ and $\Delta{}B=(\Delta{}b_1, \Delta{}b_2, \Delta{}b_3)$ are merged into the combined spike train $\Delta{}Y = (\Delta{}y_1, \Delta{}y_2, ...) = (\Delta{}a_1, \Delta{}b_1-\Delta{}y_1, ...)$.}
	\label{fig:diffspikes}
\end{figure}

\begin{proof}[Sketch of proof]
    Proving the above proposition can be done based on the observation that the first elements of differential time represented sequences are in absolute relation to each other as they can be interpreted as absolute times to a common reference time $t_0=0$.  Fig.~\ref{fig:diffspikes} illustrates this concept. An ordering with respect to their difference times is guaranteed only for the first elements of the differential time represented sequences. Therefore, the first delta time of the merged spike train $\Delta Y$ must be the minimum among the first elements of the input spike trains. Updating the other first elements (in the general case of more than two spike trains) by subtracting this minimum time can be interpreted as moving the set zero timeline (i.e. resetting the timeline to a new shifted zero). After the subtractions, all first elements of the differential time represented input spike trains are in absolute relation again, although now with the shifted set time. The sequence $\Delta Y$ is then generated by successively outputting the minimum remaining elements.
\end{proof}

\ifappendix
For a more formal proof we refer to the Appendix. Structurally similar arguments appear in the literature on stochastic processes and cyber forensics \cite{kingman-poisson-processes,stevens2004,beyond-timestamps}. 

\fi

\subsection{Merging spike trains in hardware}
\begin{figure}[b]
    \centering
    \includegraphics[width=0.7\linewidth]{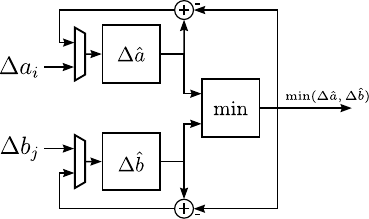}
    \caption{Spike merger element capable of merging two separate spike trains represented in differential time. The difference time of the incoming spikes is compared, with the minimum value being propagated to the output and simultaneously subtracted from the larger remaining spike. This process repeats, until no input spikes remain. The result is a single \ac{dte} sequence combining the two input sequences. Cascading this merger element into a tree with depth $\log_2(M)$ (with $M$ being the total number of input spikes) allows for the merging of an arbitrary number of spike trains.}
    \label{fig:spike_merger}
\end{figure}

This merging of the spikes can be accomplished by directly implementing Proposition~\ref{prop:merge_spikes}, leading to the spike merger element shown in \figref{fig:spike_merger}, with the variables $\Delta{}\hat{a}$ and $\Delta{}\hat{b}$ representing the continuously updated $\Delta{}a_i$ and $\Delta{}b_j$ respectively. This update is either by subtraction or by replacement with the next difference time of the spike train.

The hardware block in \figref{fig:spike_merger} merges two separate \ac{dte} spike trains, producing a new \ac{dte} spike train, containing all spikes from both input spike trains. By cascading the spike merger elements, any number of spike trains can be efficiently combined into a single spike train, using a tree structure of depth $\lceil\log_2(M)\rceil$ with $M-1$ spike merger elements, where $M$ is the number of separate spike trains entering the network.

While this merger element allows finding the spike that occurs next, it obfuscates the information of which spike train produced that spike. To recover this information, we introduce a separate vector which is extended by one bit during each stage of the $\log_2(M)$ tree. Depending on whether the upper or lower spike train produced the minimum for each spike merger element, either a $1$ or a $0$ is appended to the vector, thus producing an index identifying the originating spike train. The hardware resources required for the entire spike merger tree for each of the three evaluated datasets is listed in~\tabref{tab:res_use_spike_merger}.

\subsection{Evaluation of DTR spike merger compared to merging in absolute time}
\begin{figure}[b]
    \centering
    \includegraphics[width=0.9\linewidth]{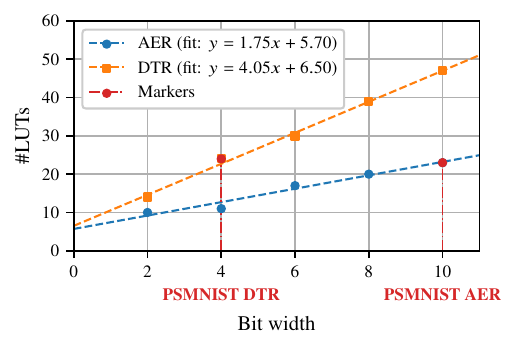}
    \caption{LUT Usage for a single comparison element (see \figref{fig:spike_merger}) used in the spike merger tree for difference time (DTR) and absolute time (AER) represented inputs. While DTR requires more resources for the same number of input bits, the actual number of input bits that are required for DTR are lower, as seen by the two markers indicating the used bit width for DTR and AER for the PSMNIST dataset.}
    \label{fig:aer_vs_dtr}
\end{figure}

Instead of directly merging the \ac{dtr} represented spikes, one might think to convert the spikes back to absolute time and sort according to absolute time. As we will show here, this approach incurs additional overhead in terms of hardware resources compared to our method. As an example, assume that we once again wish to determine the earliest spike, among a number of distinct (but sorted in time) spike trains, all given in absolute time. A simple linear search through all of the first spikes would suffice, but would result in $\mathcal{O}(n)$ time complexity for every single spike. Instead, one would quickly arrive at a tree structure, similar to the one described previously, but with comparison elements that simply forward the minimum of the two received spikes, i.e. \figref{fig:spike_merger} without the subtraction mechanism. Synthesizing both of these variants (i.e. the spike merger element for the differential time and the one for absolute time spikes) for multiple bit widths, leads to the plot in \figref{fig:aer_vs_dtr}. This plot clearly demonstrates that, for the same bit widths, the comparators used for \ac{aer} require less resources (in terms of LUTs), compared to the more complex tree elements for \ac{dte} (cf. \figref{fig:spike_merger}). Since the number of representational bits for \ac{dtr} is lower than for \ac{aer}, using the actual bit width chosen for the dataset, one can see that (for \ac{dtr}) the LUT resource usage is comparable, while the resource usage in terms of registers is $2.5$x lower. Therefore, even assuming \ac{aer} inputs and \ac{aer} outputs, locating the spike to occur next using \ac{dtr} saves resources compared to searching in absolute time.

\begin{table}[ht]
    \centering
    \renewcommand{\arraystretch}{1.3}
    \begin{threeparttable}
        \caption{Resource utilization for spike merger used in the accelerators.}
        \begin{tabular}{lccc}
         \hline
         Dataset & \#LUTs & \#FF & Power [mW]  \\
         \hline
         SHD     & 7054     & 4158     & 137\\
         SSC     & 7055     & 4158     & 156\\
         PSMNIST & N/A$^\S$ & N/A$^\S$ & N/A$^\S$\\
         \hline
    \end{tabular}
        
    \begin{tablenotes}
        \item[$\S$] Not applicable since network only has one input, so no spike merger needed. 
    \end{tablenotes}
        
    \label{tab:res_use_spike_merger}
    \end{threeparttable}
\end{table}


\section{Modular hardware architecture}
\label{sec:hw_arch}



\begin{figure}
    \centering
    \includegraphics[width=\linewidth]{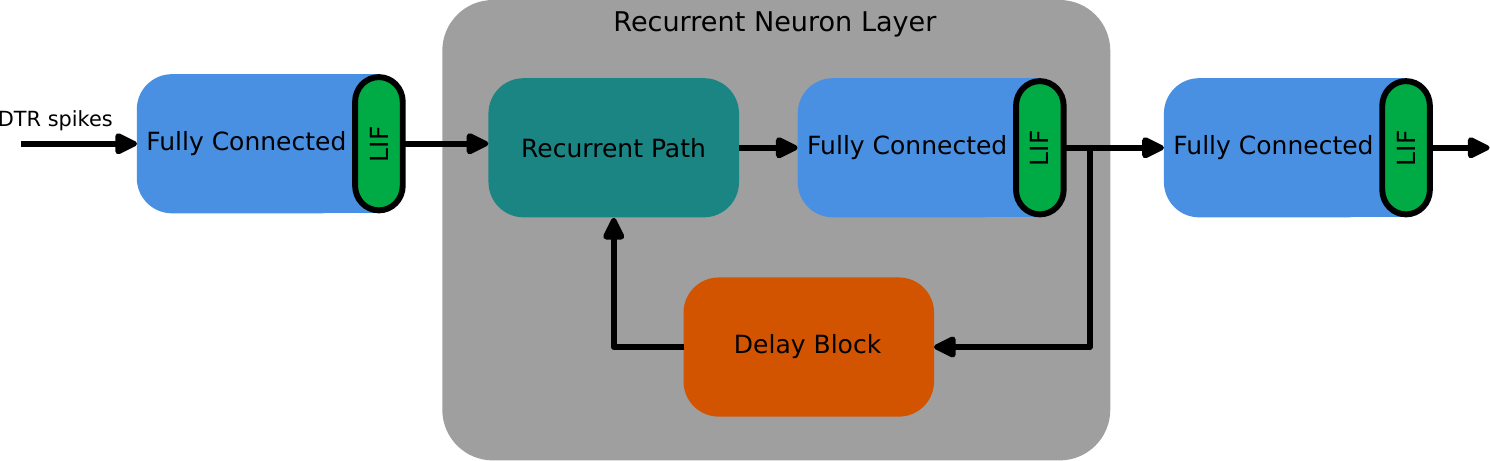}
    \caption{Hardware system architecture schematic of the accelerator used for the \ac{shd} dataset. The network consists of two feedforward (fully connected) linear layers, each with a LIF activation layer, and a recurrent neuron layer, which is also followed by a LIF activation layer. The recurrent neuron layer also includes axonal delays (i.e. the outputs from specific neurons are delayed by a specified number of time steps), which are learned alongside the weights during the training process.}
    \label{fig:shd_hw_arch}
\end{figure}

We propose a library of modular primitives, which can be combined in flexible configurations to enable building a custom accelerator which prioritizes a streaming dataflow instead of routing data to separate computational tiles. The hardware blocks include:
\begin{description}
    \item[Fully connected block] Implements a standard linear layer, followed by a neuron activation layer.
    \item[Recurrent path block] Connects the outputs of a fully connected block back to the inputs, thus enabling recurrent neuron layers.
    \item[Delay block] Enables delaying spikes by an arbitrary amount of cycles, allowing for arbitrary learned axonal delays in the recurrent and forward path.   
\end{description}
These blocks can be flexibly interconnected, supporting even parallel paths to enable the construction of residual connections, similar to those found in ResNet architectures\cite{he2015deepresiduallearningimage}. An example network configuration, employed for the \ac{shd} dataset, is shown schematically in \figref{fig:shd_hw_arch}. It consists of two layers of fully connected blocks with \ac{lif} activations, as well as an intermediate layer including a recurrent path and axonal delays defined during the training process. 

To enable a flexible connection of these blocks, the AXI4-Stream interface\cite{arm_amba_axi4_stream} with backpressure was used. Using the \emph{valid-ready} handshake defined in the AXI4-Stream protocol allows the different hardware blocks to work independently of each other, without losing data in the process. Furthermore, well-defined interfaces for the blocks also enable them to be developed in isolation, improving validation and development. 

The \ac{dte} spikes used for the hardware consist of an axonal index, identifying the neuron in the previous layer which produced this spike, as well as the differential time indicating the number of integer time steps since the last spike. The inputs to the first layer of the network are assumed to be from input encoding layers placed before the network. All blocks in the network receive, process and produce single spikes, \ie all of the spikes in the network are processed sequentially. While this does impact performance compared to processing multiple spikes in parallel, it significantly simplifies the hardware and allows maximum utilization of sparse spike trains, \ie time steps at which no spike occurs are skipped. Furthermore, processing multiple spikes in parallel does not provide significant benefits for spikes occurring at different time steps, due to the nonlinear thresholding operation of the neuron incurring a temporal dependence between spikes.

\subsection{Fully Connected Block}
The fully connected block, illustrated in \figref{fig:shd_hw_arch}, is the basic building block of these architectures. It represents the standard matrix-vector multiplication 
\begin{equation}
y = \mathbf{W}x,
\label{eq:matrix_vector_product}    
\end{equation}
where $\mathbf{W}\in\mathbb{R}^{n\times{}m}$ is a matrix and $y\in\mathbb{R}^n$ and $x\in\{0, 1\}^m$ are vectors. Specifically, $\mathbf{W}$ is the matrix of weights learned during the training process of the network and $x$ is the input vector consisting of spikes. Due to 0/1 multiplicands, this multiplication results in an accumulative process in the hardware, \ie upon receiving an event from axon i, the block retrieves the i-th column of the weight matrix $\mathbf{W}$ and updates the corresponding neuron potentials. After the matrix-vector multiplication, a \ac{lif} activation layer is applied, the dynamics of which are as follows:

\begin{align}
    \begin{split}
        P_{k} &= P_{k-1}\beta^{\Delta t} + y_k - r_{out}\\
        r_{out} &= \begin{cases}
        \mbox{$\theta$,} & \mbox{if } P_{k} \ge \theta\\
        \mbox{0,} & \mbox{if } P_{k} < \theta,
        \end{cases}
    \end{split}
    \label{eq:lif_neuron}
\end{align}
where $P_k\in\mathbb{R}$ is the neuron potential at time step $k$, $\beta=0.5$ is the decay rate of the potential, $\Delta{}t\in\mathbb{N}_0$ is the elapsed time since the last event, $k\in\mathbb{N}_0$ is the time at which at least one input spike $s_i \in \{0, 1\}$ occurred, $y_k$ is the value resulting from the contribution of the weight matrix $\mathbf{W}$ and the input spike vector $x_k$ at time step $k$, $r_{out}\in\mathbb{R}$ represents the reset value and $\theta\in\mathbb{R}$ is the threshold. Notably, the decay of the neuron potential requires the delta time since the last spike, which directly leverages the \acf{dte}. The primary function of the fully connected block is to realize the matrix-vector product from \eqref{eq:matrix_vector_product} and the activation rule based on \eqref{eq:lif_neuron}. Since the input to the block consists of single spikes, neither of these equations requires multiplications. Specifically, by choosing $\beta=0.5$, the potential decay can be implemented as a right-shift, thereby eliminating the need for costly DSP slices. 

\begin{figure}
    \centering
    \includegraphics[width=0.8\linewidth]{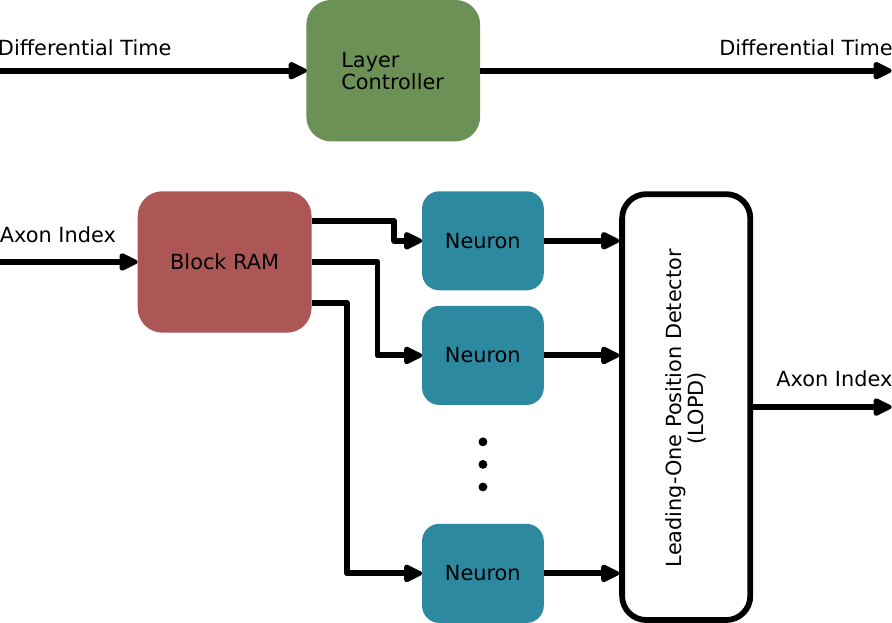}
    \caption{Internal structure of the fully connected block, implementing a linear layer, followed by a neuron activation such as \ac{lif}. The axonal index is directly used as address for the BRAM, which stores the weights of the network. This weight vector is provided to the respective neuron processing elements, which then handle the neuron dynamics and spike generation. The spikes are fed into the \ac{lopd}, ensuring that the currently processed spike is reset after it is passed to the next layer. The layer controller is responsible for orchestrating the dataflow as well as sending the decay signals to the individual neurons, triggering the potential decay process.}
    \label{fig:fully_connected_arch}
\end{figure}

A block diagram of the entire fully connected unit is shown schematically in \figref{fig:fully_connected_arch}, with full resource usage for each layer of each neural network after synthesis listed in \tabref{tab:res_use_fully_connected}. Each incoming spike consists of the differential time and the axonal index, which are processed individually as shown in \figref{fig:fully_connected_arch}. The axonal index can be used directly as an address to access the BRAM containing the weights that are subsequently added to the individual neuron potentials. If the number of neurons is large, multiple block RAMs are instantiated in parallel to ensure sufficient bandwidth for retrieving the weights. 

The neurons that receive these weights directly from the BRAM are standard LIF neurons, which consist of a register for storing the neuron potential coupled with logic to detect the threshold crossing and implement the potential decay. While only LIF neurons are used here, the modular design and well-defined interfaces between components allow for seamless replacement of the neuron with a different implementation, such as the more biologically plausible Izhikevich model \cite{Izhikevich}.

\begin{table}[ht]
  \centering
  \caption{FPGA resource utilization and power of the fully connected
    layers on the ZCU102 (XCZU9EG) with no time-multiplexing. Percentages are relative to the
    device totals (274\,080 LUTs, 548\,160 FFs, 32\,832\,Kbit BRAM).}
  \label{tab:res_use_fully_connected}
  \setlength{\tabcolsep}{4pt}
  \begin{tabular}{
      c
      S[table-format=2.1] S[table-format=1.1]
      S[table-format=1.1] S[table-format=1.1]
      S[table-format=4.0] S[table-format=1.1]
      S[table-format=3.0]
    }
    \toprule
    Layer
      & \multicolumn{2}{c}{LUTs}
      & \multicolumn{2}{c}{FFs}
      & \multicolumn{2}{c}{BRAM}
      & {Power} \\
    \cmidrule(lr){2-3} \cmidrule(lr){4-5} \cmidrule(lr){6-7}
      & {[k]} & {[\%]}
      & {[k]} & {[\%]}
      & {[Kbit]} & {[\%]}
      & {[mW]} \\
    \midrule
    \multicolumn{8}{c}{SHD} \\
    \midrule
    1 & 24.9 & 9.1 & 6.8 & 1.2 &  315 & 1.0 & 799 \\
    2 & 24.9 & 9.1 & 6.8 & 1.2 & 1152 & 3.5 & 682 \\
    3 &  1.5 & 0.6 & 0.6 & 0.1 &   45 & 0.1 &  74 \\
    \midrule
    \multicolumn{8}{c}{SSC} \\
    \midrule
    1 & 23.8 & 8.7 & 7.3 & 1.3 &  792 & 2.4 & 622 \\
    2 & 23.8 & 8.7 & 7.3 & 1.3 & 1024 & 3.1 & 876 \\
    3 & 23.8 & 8.7 & 7.3 & 1.3 & 1024 & 3.1 & 907 \\
    4 &  3.0 & 1.1 & 1.1 & 0.2 &   70 & 0.2 & 136 \\
    \midrule
    \multicolumn{8}{c}{PSMNIST} \\
    \midrule
    1 &  4.2 & 1.5 & 1.6 & 0.3 &   37 & 0.1 & 116 \\
    2 & 14.5 & 5.3 & 4.8 & 0.9 &  457 & 1.4 & 554 \\
    3 & 14.1 & 5.1 & 4.8 & 0.9 &  702 & 2.1 & 576 \\
    4 &  0.7 & 0.2 & 0.3 & 0.1 &   17 & 0.1 &  29 \\
    \bottomrule
  \end{tabular}
\end{table}

\subsection{Dynamic neuron reuse}
\begin{figure}
    \centering
    \includegraphics[width=0.8\linewidth]{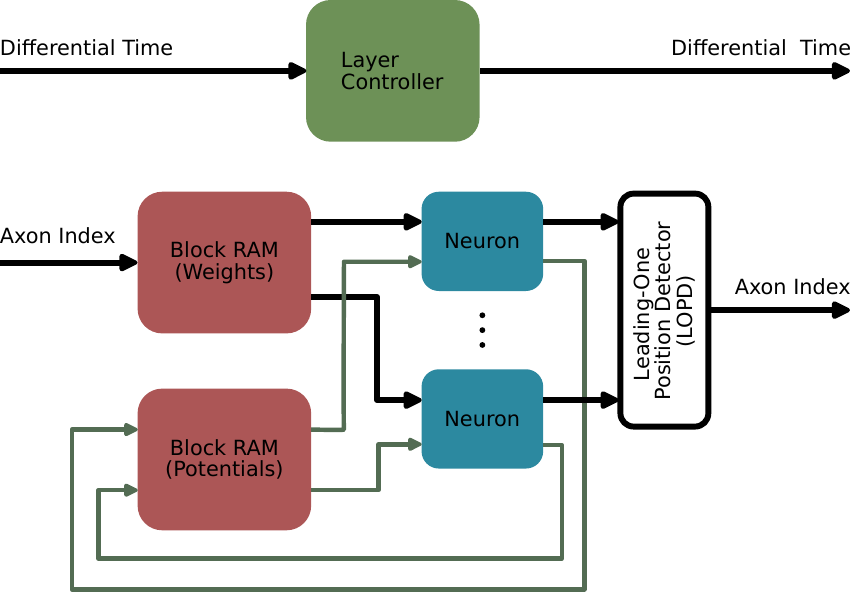}
    \caption{Internal structure of the fully connected block when using dynamic neuron reuse with a limited number of neurons (\ie fewer than present in the neural network). The weights are retrieved by indexing into the BRAM directly using the axonal index, while another BRAM stores the corresponding neuron potentials. Since only a subset of the required number of neurons is present in hardware, the potentials must also be dynamically reloaded for each batch of neurons being processed. The \ac{lopd} is responsible for serializing the spikes for the next layer. The transition to the subsequent batch of neurons, weights, and potentials occurs once the \ac{lopd} has processed all current spikes.}
    \label{fig:fully_connected_arch_dynamic_reload}
\end{figure}

Since the hardware resources of the architecture naturally scale with the number of neurons and weights, the requirement to instantiate all neurons and their respective neuron potentials results in substantial LUT and flip-flop utilization. However, a tradeoff between performance and required hardware can be implemented, by time-multiplexing the existing neurons. Instead of maintaining all neuron potentials in registers concurrently, and thus requiring weights for all neurons to be loaded in a single clock cycle, we can instantiate a smaller number of neurons (i.e. \quote{physical} neurons) in the design, and successively load the required potentials for these neurons. This, however, necessitates a mechanism to load and store the respective neuron potentials. An example of this strategy is shown schematically in \figref{fig:fully_connected_arch_dynamic_reload}. Using this time-multiplexed fully connected block with dynamic neuron reuse, the required hardware is significantly decreased if the overall throughput of spikes can be slightly lowered. 


When one wants to reduce the number of instantiated neurons, a natural question is the number of neurons to choose per layer. The hardware shown in \figref{fig:fully_connected_arch_dynamic_reload} computes the logical neurons of a single layer in batches, i.e. instead of instantiating all neurons, only a smaller subset of physical neurons are actually present in the hardware which are then time-multiplexed to provide functionality equivalent to a fully instantiated layer. This dynamic reloading only takes one clock cycle (as the writing and reading of the neuron potential for all neurons may be done at the same time), while the processing of the decay, activation and adding the weights to the neuron potential has remained the same as for the fully instantiated layer and is independent of the number of instantiated neurons. Therefore, the number of clock cycles required to compute the entire network must depend on the number of batches each of the layers requires. Since the layers are connected to each other directly without FIFOs or buffers in between, they can stall each other if not enough data is available. Furthermore, if no stalling or blocking occurs, all of the layers are capable of working in parallel. Both of these combined makes an even distribution of batches, i.e each layer has an equal number of batches to compute, a natural choice. We term this assignment proportional, as the number of physical neurons per layer is directly proportional to the number of neurons present in the layer. 

\begin{figure}
    \centering
    \includegraphics[width=0.85\linewidth]{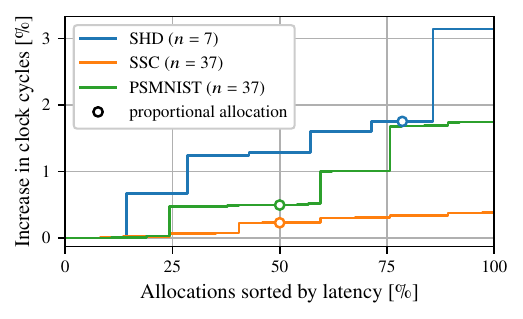}
    \caption{Relative increase in total clock cycles, in percent, for all evaluated
allocations of the fixed budget of 276 physical neurons. Each value is
measured with respect to the fastest allocation of the respective
dataset. Within each dataset, allocations are enumerated on the $x$-axis
from fastest to slowest, with one data point per allocation. $n$ denotes the number of allocations evaluated per dataset.
Circles mark the allocation proportional to the logical layer sizes.}
    \label{fig:total_cycles_normalized}
\end{figure}

\begin{figure}
    \centering
    \includegraphics[width=0.8\linewidth]{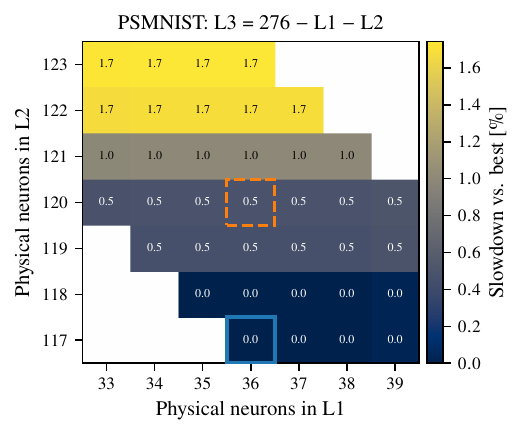}
    \caption{Latency of the PSMNIST network vs.\ the allocation of physical neurons.
The three dynamic reuse recurrent layers L1--L3 of the
(1--64--212--212--10) network share a fixed budget of 276 physical
neurons. The last layer is fully instantiated. Each cell
shows the percentage increase in total clock cycles relative to the
fastest allocation, averaged over 40 test samples of 784 time steps, with
$\mathrm{L3}=276-\mathrm{L1}-\mathrm{L2}$. Solid and dashed outlines mark
the fastest allocation (36/117/123) and the allocation proportional to
the logical layer sizes (36/120/120), respectively. White cells were not
simulated because L3 would fall outside the evaluated range of 117--123
physical neurons.}
    \label{fig:psmnist_heatmap}
\end{figure}

While this proportional assignment is not guaranteed to be optimal across all datasets, neuron configurations and hyperparameters, simulations indicate that it is very close to the best allocation possible. \figref{fig:total_cycles_normalized} shows that the best allocation is only 1-2 percentage points removed from the proportional assignment when tested in a neighborhood of three (all variations of neuron assignments up to three above and three below the proportional) across all three datasets, averaged over 40 random runs. The results indicate that this proximity to the optimal allocation also holds for allocations that are further removed from the proportional assignment, as shown in \figref{fig:psmnist_heatmap}. Furthermore, \figref{fig:psmnist_heatmap} also shows that the this landscape of allocations seems to be approximately convex around the optimal point, which the proportional assignment is near, thus also enabling a process of iterative improvement of this assignment towards the optimum.

For applications seeking to minimize weight data transfers, a weight-stationary approach such as the one introduced in \cite{10.1145/3510854} may be used, which increases the reuse of weights by utilizing a time window size optimization technique (TWSO). Using TWSO the weights necessary for computing the feedforward path may be reused for multiple time steps, thus directly decreasing the number of RAM transfers. However, this technique necessitates a buffer for storing multiple time steps of feedforward spikes, which is why it is not utilized here.

\subsection{Recurrent Layer \& Recurrent Path Block}
A recurrent layer can be constructed by extending the fully connected block. As mentioned before, the fully connected block computes the equation $\mathbf{y} = \mathbf{W}\mathbf{x}$ where $\mathbf{W}\in\mathbb{R}^{n\times{}m}$, which is similar to the equation used for recurrent fully connected layers
\begin{equation}
    \mathbf{y} = \mathbf{W}\mathbf{x} + \mathbf{W_r}\mathbf{x_r}
\end{equation}
where $\mathbf{W_r}\in\mathbb{R}^{n\times{}n}$ and $x_r\in\{0,1\}^n$ are the recurrent weights and recurrent inputs respectively. From these, we can then construct a new matrix of weights
\begin{equation}
    \mathbf{W'} = \left[\begin{array}{cc}
        \mathbf{W} & \mathbf{W}_r 
    \end{array}\right]\in\mathbb{R}^{n\times{}(m+n)}
\end{equation}
along with a new vector of the combined inputs
\begin{equation}
    \mathbf{x'} = \left[\begin{array}{c}
        \mathbf{x}\\
        \mathbf{x_r}
    \end{array}\right]\in\{0,1\}^{m+n}.
\end{equation}
Rewriting \eqref{eq:matrix_vector_product} with these, we then get 
\begin{equation}
    \mathbf{y} = \mathbf{W'}\mathbf{x'} = \left[\begin{array}{cc}
        \mathbf{W} & \mathbf{W}_r
    \end{array}\right]
    \left[\begin{array}{c}
        \mathbf{x}\\
        \mathbf{x_r}
    \end{array}\right]
\end{equation}
which can be used with the existing fully connected block. The weight matrices $\mathbf{W}$ and $\mathbf{W_r}$ can be combined into the matrix $\mathbf{W'}$ before the design is synthesized. The inputs $\mathbf{x}$ and $\mathbf{x_r}$, however, need to be merged at runtime, for which the recurrent path block is responsible, which acts as an arbiter.

In the hardware, the two inputs $\mathbf{x}$ and $\mathbf{x_r}$ are streams of \ac{dte} spikes, consisting of both the spikes originating from the previous layer, as well as the recurrent spikes. Initially, all spikes from the previous layer enter the recurrent path block until the last spike of that time step initiates the spike generation from the fully connected block. These spikes then enter the recurrent path block, which processes all the recurrent spikes until this time step is also complete, at which point the recurrent path block will resume processing the spikes of the previous layer.

\begin{figure}
    \centering
    \includegraphics[width=0.8\linewidth]{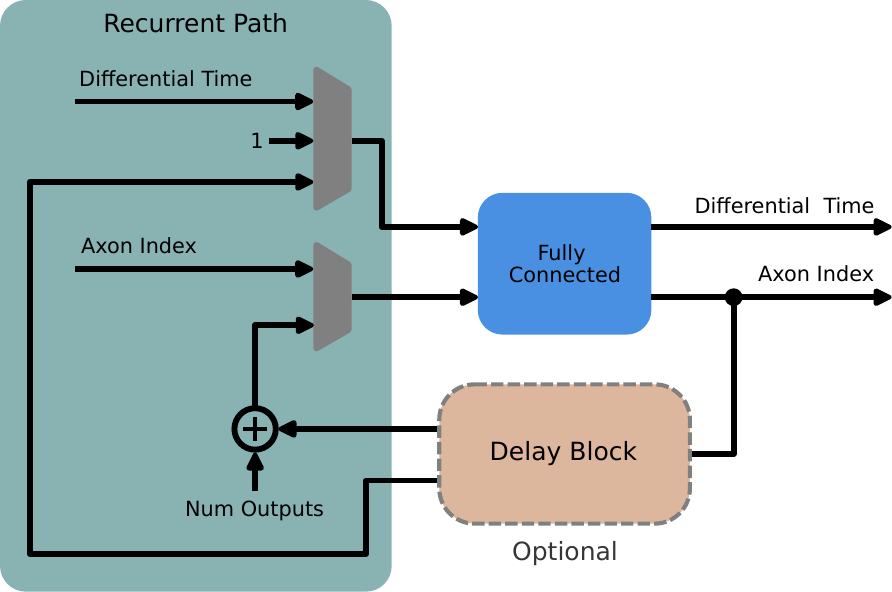}
    \caption{The recurrent layer is constructed by reusing the fully connected block. The recurrent path block either forwards the incoming spikes (from the previous layer) or the recurrent spikes generated by this layer. The axonal index is adjusted by adding the number of outputs of the current layer to ensure correct weight selection in the fully connected block. The differential time is determined by the delay block, which may be added to incorporate longer recurrent axonal delays to the layer. For directly recurrent spikes, i.e., spikes without axonal delay, this time is set to 1.}
    \label{fig:recurrent_layer}
\end{figure}

To ensure that the recurrent spikes ($\mathbf{x_r}$) select the correct weights ($\mathbf{W_r}$), their axonal index must be adjusted. This is accomplished by adding the number of inputs to the axonal index of the recurrent spike, \ie adding the offset at which the recurrent weights are positioned in memory. Furthermore, since recurrent spikes, in their simplest form, occur exactly 1 time step after the spike which triggered them, the time difference of the recurrent spike is set to $1$, assuming no axonal delays are used. In the case of learned axonal delays, the time difference of the recurrent spike is directly determined by the delay block. A schematic block diagram showing a recurrent layer including the recurrent path block is illustrated in \figref{fig:recurrent_layer} with the full resource usage after synthesis being listed in \tabref{tab:res_use_rec_layer}.

\begin{table}[ht]
  \centering
  \caption{FPGA resource utilization and power of the recurrent layers
    on the ZCU102 (XCZU9EG) with no time-multiplexing. Percentages are relative to the device
    totals (274\,080 LUTs, 548\,160 FFs, 32\,832\,Kbit BRAM).
    Non-recurrent layers are omitted.}
  \label{tab:res_use_rec_layer}
  \setlength{\tabcolsep}{4pt}
  \begin{tabular}{
      c
      S[table-format=2.1] S[table-format=1.1]
      S[table-format=1.1] S[table-format=1.1]
      S[table-format=4.0] S[table-format=1.1]
      S[table-format=3.0]
    }
    \toprule
    Layer
      & \multicolumn{2}{c}{LUTs}
      & \multicolumn{2}{c}{FFs}
      & \multicolumn{2}{c}{BRAM}
      & {Power} \\
    \cmidrule(lr){2-3} \cmidrule(lr){4-5} \cmidrule(lr){6-7}
      & {[k]} & {[\%]}
      & {[k]} & {[\%]}
      & {[Kbit]} & {[\%]}
      & {[mW]} \\
    \midrule
    \multicolumn{8}{c}{SHD} \\
    \midrule
    2 & 25.2 & 9.2 & 7.0 & 1.3 & 1441 & 4.4 & 704 \\
    \midrule
    \multicolumn{8}{c}{SSC} \\
    \midrule
    1 & 24.1 & 8.8 & 7.6 & 1.4 & 1081 & 3.3 & 638 \\
    2 & 24.1 & 8.8 & 7.6 & 1.4 & 1313 & 4.0 & 892 \\
    3 & 24.1 & 8.8 & 7.6 & 1.4 & 1212 & 3.7 & 926 \\
    \midrule
    \multicolumn{8}{c}{PSMNIST} \\
    \midrule
    1 &  4.4 & 1.6 & 1.8 & 0.3 &  166 & 0.5 & 125 \\
    2 & 14.8 & 5.4 & 5.1 & 0.9 &  746 & 2.3 & 570 \\
    3 & 14.4 & 5.3 & 5.1 & 0.9 &  991 & 3.0 & 595 \\
    \bottomrule
  \end{tabular}
\end{table}

\subsection{Delay Block}
\begin{figure}[b]
    \centering
    \includegraphics[width=0.8\linewidth]{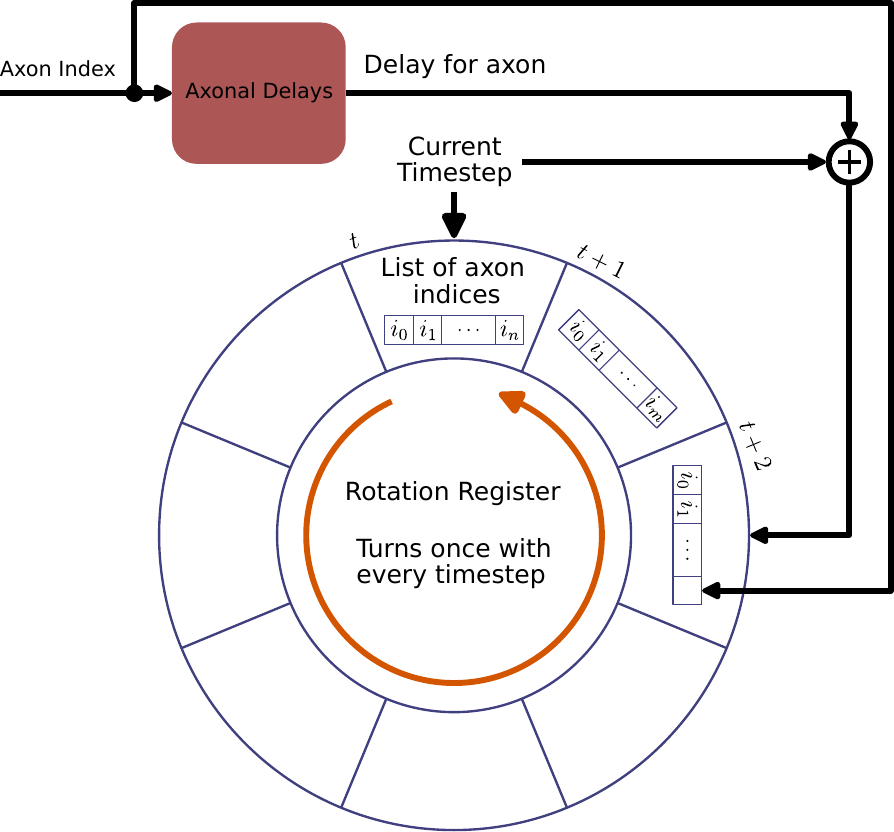}
    \caption{Schematic depiction of rotational memory inside the delay block being used to store and retrieve spikes. The axon index is used to retrieve the previously stored delay which, in conjunction with the offset pointer, is used to determine the address at which to place the spike for later retrieval. The rotational memory is placed in a BRAM to ensure efficient access. The spikes stored in the location designated by the offset pointer are sent out every time step, after which the offset pointer advances.}
    \label{fig:delay_block}
\end{figure}
The delay block implements a programmable latency by delaying the incoming spikes for a specified number of time steps, which can be configured per axonal index, and outputs the previously received spikes at the correct time step. Therefore, a mechanism is required to retrieve the stored delay for a spike of a given axonal index and to store incoming spikes when their delay is larger than 0.

The delay block, like the fully connected block, also receives a single spike as input and produces a single spike as output. This integrates seamlessly into \ac{dte}-based architectures, as axonal delays are naturally differentially time-represented relative to the triggering spike. Because the hardware module retrieves the stored delay based on the axonal index of the spike, the specific differential time of the incoming spike is not required for this operation.

The processing stages for an input spike in the delay block are as follows:
\begin{enumerate}
    \item First, the axonal index is used to address the delay memory, \ie the BRAM is accessed to retrieve the delay assigned to the neuron that produced this spike.
    \item If the delay is $0$, the spike is forwarded to the output stage directly.
    \item If the delay is greater than zero, the spike is delayed by that number of time steps. To enable this, it is stored in a local BRAM to be retrieved later. To simplify retrieval, the spike is stored at a location in RAM that is based on its delay, \ie it is aggregated into temporal bins of spikes that occur at the same time step. This is accomplished efficiently by storing the bins in a rotational memory, as seen in \figref{fig:delay_block}. Spikes are grouped based on which time step in the future the delay block must forward them, by adding the delay to the current offset pointer. The entire rotational memory can then be placed in BRAM to ensure efficient indexing and access.
\end{enumerate}

The offset pointer is incremented with every time step of the incoming spikes. Therefore, it always points to the bin containing the spikes that occur in the current time step. Consequently, this offset pointer is used to retrieve the spikes that have been previously stored and are due for emission in the current cycle.  

The full resource usage for the delay block for each layer and each neural network after synthesis is listed in \tabref{tab:res_use_delay_block}.

\begin{table}[!t]
  \centering
  \caption{FPGA resource utilization and power of the delay block
    for each recurrent layer on the ZCU102 (XCZU9EG) with no time-multiplexing. Percentages are
    relative to the device totals (274\,080 LUTs, 548\,160 FFs,
    32\,832\,Kbit BRAM). Non-recurrent layers are omitted.}
  \label{tab:res_use_delay_block}
  \setlength{\tabcolsep}{4pt}
  \begin{tabular}{
      c
      S[table-format=3.0] S[table-format=1.2]
      S[table-format=3.0] S[table-format=1.2]
      S[table-format=3.0] S[table-format=1.2]
      S[table-format=2.0]
    }
    \toprule
    Layer
      & \multicolumn{2}{c}{LUTs}
      & \multicolumn{2}{c}{FFs}
      & \multicolumn{2}{c}{BRAM}
      & {Power} \\
    \cmidrule(lr){2-3} \cmidrule(lr){4-5} \cmidrule(lr){6-7}
      & {[\#]} & {[\%]}
      & {[\#]} & {[\%]}
      & {[Kbit]} & {[\%]}
      & {[mW]} \\
    \midrule
    \multicolumn{8}{c}{SHD} \\
    \midrule
    2 & 220 & 0.08 & 168 & 0.03 & 289 & 0.88 & 14 \\
    \midrule
    \multicolumn{8}{c}{SSC} \\
    \midrule
    1 & 245 & 0.09 & 168 & 0.03 & 289 & 0.88 & 15 \\
    2 & 219 & 0.08 & 168 & 0.03 & 289 & 0.88 & 13 \\
    3 & 217 & 0.08 & 168 & 0.03 & 289 & 0.88 & 14 \\
    \midrule
    \multicolumn{8}{c}{PSMNIST} \\
    \midrule
    1 & 148 & 0.05 & 148 & 0.03 & 129 & 0.39 &  8 \\
    2 & 235 & 0.09 & 168 & 0.03 & 289 & 0.88 & 15 \\
    3 & 234 & 0.09 & 168 & 0.03 & 289 & 0.88 & 17 \\
    \bottomrule
  \end{tabular}
\end{table}

\section{Reproducibility \& Code Availability}
\label{sec:reprod}

To evaluate our hardware against other existing accelerators, three standard neuromorphic benchmark datasets were used for comparison. These are \acf{shd}, \acf{ssc} and \acf{psmnist}. \ac{shd} and \ac{ssc} are audio-based datasets and \ac{psmnist} is an image-based dataset, demonstrating that the \acp{snn} used here perform well on short audio datasets as well as long-range sequential patterns. As \ac{shd} and \ac{ssc} are established neuromorphic benchmark datasets, they already directly provide spikes and spike trains, i.e. no coding such as Rate Coding or Latency Coding is necessary. The \ac{psmnist} dataset contains raw pixel data that was thresholded to binarize the inputs.

Three different \acp{snn} were trained for the datasets, with the initial configuration and hyperparameters of the networks adapted from \cite{queant2025delreclearningdelaysrecurrent}. The structure and layer sizes of these networks can be seen in \tabref{tab:networks}. The developed Python framework automatically generates the corresponding hardware for these architectures and can likewise be applied to other network structures.

Minor changes were applied to the networks ensure compatibility with the hardware described in this work. Specifically, the bias parameters and modified batch normalizations were omitted from the network (given that the fully connected block is designed without bias parameters, the batch normalization cannot be fused into the weights of the linear layer). Furthermore, the network for the \ac{psmnist} dataset used by \cite{queant2025delreclearningdelaysrecurrent} feeds raw pixel data into the first linear layer, requiring multiplication operations. Therefore, a thresholding parameter $0\leq\theta\leq1$ was introduced during training which was used to binarize the pixel values into the set $\{0, 1\}$. This thresholding reduces the overall accuracy by approximately $0.8\%$, yet completely eliminates the need for multiplications in hardware, and was thus deemed adequate for this use case. While this thresholding is sufficient for the \ac{psmnist} dataset, other more varied datasets might incur significant accuracy drops when using this approach. Therefore, other techniques, such as the stochastic binarization method introduced in \cite{8731879} for \acp{bnn} can be employed to further improve performance.

The training of the networks was completed using the openly available source code by \cite{queant2025delreclearningdelaysrecurrent}, using their suggested hyperparameters. The results mentioned in this work were obtained by training each of the three networks five separate times and averaging their results. For the hardware results, the best run for each dataset was quantized and the hardware for each specific network was evaluated using custom test benches. These test benches load the true quantized weights and a single randomly sampled test sample from each dataset, run it through the accelerator and subsequently compare the result to the Python result. For each accelerator, 1000 such simulations were performed. The reported hardware latency is the mean over these runs, and the hardware accuracy is the fraction of samples classified correctly.

The entire hardware design was developed in VHDL and validated alongside a bit-accurate PyTorch implementation of the networks. The RTL source code for the hardware as well as the Python-based generation tool can be found at \url{https://github.com/DanielWindhager/streaming-snn}. This repository also contains the JSON configuration files for the accelerators for the abovementioned datasets.

\begin{table}
    \centering
    \renewcommand{\arraystretch}{1.3}
    \begin{threeparttable}
        \caption{Network size and number of neurons instantiated in hardware
                 for the accelerators targeting the SHD, SSC and PSMNIST datasets.}
        \label{tab:networks}
        \begin{tabular}{lccccc}
         \hline
         Accelerator & In & Hidden 1 & Hidden 2 & Hidden 3 & Out \\
         \hline
         SHD (Par.)                & $140^\S$ & $256$               & $256^\mathrm{R}$       & --                     & $20$ \\
         SHD (TM)\tnote{*}         & $140^\S$ & $64$ (256)          & $64^\mathrm{R}$ (256)  & --                     & $20$ \\
         SSC (Par.)                & $140^\S$ & $256^\mathrm{R}$    & $256^\mathrm{R}$       & $256^\mathrm{R}$       & $35^\ddag$ \\
         SSC (TM)\tnote{*}         & $140^\S$ & $64^\mathrm{R}$ (256) & $64^\mathrm{R}$ (256) & $64^\mathrm{R}$ (256) & $35^\ddag$ \\
         PSMNIST (Par.)            & $1$      & $64^\mathrm{R}$     & $212^\mathrm{R}$       & $212^\mathrm{R}$       & $10^\ddag$ \\
         PSMNIST (TM)\tnote{*}     & $1$      & $32^\mathrm{R}$ (64) & $64^\mathrm{R}$ (212) & $64^\mathrm{R}$ (212)  & $10^\ddag$ \\
         \hline
        \end{tabular}
        \begin{tablenotes}
            \item[$*$] Time-multiplexed (TM) using dynamic neuron reuse: same network as the parallel (Par.)
                       version. Values give the physical neurons instantiated in
                       hardware, with the logical neurons they emulate in parentheses.
            \item[$\S$] Inputs are reduced from the original 700 inputs by binning
                        with a factor of 5.
            \item[R] Recurrent layer with axonal delays in the recurrent path.
            \item[$\ddag$] Outputs from the last linear layer are used directly,
                           without LIF activation.
        \end{tablenotes}
    \end{threeparttable}
\end{table}

\section{Experimental Results}
\label{sec:results}

\bgroup
\def\arraystretch{1.3}
\begin{table*}[t]
	\centering
    \makebox[\textwidth][c]{
    \begin{threeparttable}
	\caption{Comparison to other hardware accelerator architectures.}
	\begin{tabular}{lrrrrrrrr}
        \multicolumn{9}{c}{\textbf{ASIC Implementations}} \\
        Work              & Process              & Dataset      & Accuracy [\%]             & Inf./s                    & Freq. [MHz]       & Area [mm²]                        & \#Mult.        & Power [W]      \\
        \midrule
        Ours (Par.)       & ASAP7                & \ac{shd}     & $\mathbf{92.46\pm{}0.69}$ & $\mathbf{\approx{}12027}$ & 642               & $^\ddagger$0.1073 (0.49)          & \textbf{0}     & 0.245          \\
        Ours (TM)\tnote{*}& ASAP7                & \ac{shd}     & $\mathbf{92.46\pm{}0.69}$ & $\approx{}6768         $  & \textbf{1103}     & $^\ddagger$\textbf{0.0933 (0.49)} & \textbf{0}     & 0.296          \\
        \cite{meijer2025} & GF 22nm              & \ac{shd}     & 87                        & $\approx235$              & Not reported      & 1.45                              & Not reported   & Not reported   \\
        \cite{10558209}   & Intel 14nm (Loihi)   & \ac{shd}     & $89.04$                   & $109.89$                  & Not reported      & Not applicable                    & Not applicable & \textbf{0.004} \\
        Ours (Par.)       & ASAP7                & \ac{ssc}     & $\mathbf{82.08\pm{}0.18}$ & $\mathbf{\approx{}7032}$  & 822               & $^\ddagger$0.2033 (0.49)          & \textbf{0}     & 0.506          \\
        Ours (TM)\tnote{*}& ASAP7                & \ac{ssc}     & $\mathbf{82.08\pm{}0.18}$ & $\approx{}2752         $  & \textbf{1069}     & $^\ddagger$\textbf{0.1789 (0.49)} & \textbf{0}     & \textbf{0.441} \\
        Ours (Par.)       & ASAP7                & \ac{psmnist} & $\mathbf{94.86\pm{}0.24}$ & $\mathbf{\approx{}11937}$ & 985               & $^\ddagger$\textbf{0.0879 (0.49)} & \textbf{0}     & 0.233          \\
        Ours (TM)\tnote{*}& ASAP7                & \ac{psmnist} & $\mathbf{94.86\pm{}0.24}$ & $\approx{}6223         $  & \textbf{1119}     & $^\ddagger$0.0885 (0.49)          & \textbf{0}     & \textbf{0.222} \\\\
        

        \multicolumn{9}{c}{\textbf{FPGA Implementations}} \\
        Work                  & FPGA            & Dataset      & Accuracy [\%]             & Inf./s                   & Freq. [MHz]       & \#LUTs            & \#DSP          & Power [W]          \\
        \midrule
        Ours (Par.)           & ZCU102          & \ac{shd}     & $92.46\pm{}0.69$          & $\mathbf{\approx{6023}}$ & 328               & 53706             & \textbf{0}     & 3.325              \\
        Ours (TM)\tnote{*}    & ZCU102          & \ac{shd}     & $92.46\pm{}0.69$          & $\approx{2056}$          & \textbf{335}      & \textbf{20488}    & \textbf{0}     & 1.647              \\
        \cite{11327474}       & VCU128          & \ac{shd}     & $\mathbf{94.6}$           & Not reported             & Not reported      & $^\dagger212900$  & Not reported   & $^\P10.24$       \\
        \cite{quantisenc}     & Zynq UltraScale & \ac{shd}     & $87.8$                    & Not reported             & Not reported      & $230400$          & 1728           & 3.349              \\
        \cite{nakano_blachut} & ZCU104          & \ac{shd}     & $92.30$                   & 5586.59                  & 200               & 81567             & 318            & $3.94$             \\
        \cite{nakano_blachut} & ZCU104          & \ac{shd}     & $88.78$                   & 5714.29                  & 200               & 34474             & 106            & $3.79$             \\
        \cite{spiker+}        & XA7Z020         & \ac{shd}     & $72.99$                   & 1851.85                  & 100               & Not reported      & Not reported   & $\mathbf{^\S0.43}$ \\
        Ours (Par.)           & ZCU102          & \ac{ssc}     & $\mathbf{82.08\pm{}0.18}$ & $\mathbf{\approx{2644}}$ & 309               & 86636             & \textbf{0}     & 6.013              \\
        Ours (TM)\tnote{*}    & ZCU102          & \ac{ssc}     & $\mathbf{82.08\pm{}0.18}$ & $\approx{857}$           & \textbf{333}      & \textbf{31639}    & \textbf{0}     & \textbf{2.538}     \\
        \cite{11327474}       & VCU128          & \ac{ssc}     & $65.6$                    & Not reported             & Not reported      & $^\dagger338700$  & Not reported   & $^\P10.24$       \\
        Ours (Par.)           & ZCU102          & \ac{psmnist} & $\mathbf{94.86\pm{}0.24}$ & $\mathbf{\approx{4049}}$ & 334               & 37571             & \textbf{0}     & 2.936              \\
        Ours (TM)\tnote{*}    & ZCU102          & \ac{psmnist} & $\mathbf{94.86\pm{}0.24}$ & $\approx{1930}$          & \textbf{347}      & \textbf{14150}    & \textbf{0}     & \textbf{1.646}     \\

    \end{tabular}
	\label{tab:perf_comparison}
    \begin{tablenotes}
        \small
        \item[$*$] Time-multiplexed (TM) using dynamic neuron reuse.
        \item[$\ddagger$] Figure denotes utilized area. Value in parenthesis denotes overall used area including unused silicon. 
        \item[$\dagger$] Originally reported as CLBs (which would exceed the resources of the FPGA).
        \item[$\P$] Power usage higher due to external components being used.
        \item[$\S$] Although not explicitly stated, this is most likely only dynamic power since the static power draw of the FPGA alone is\\ comparable to this figure. 
        \end{tablenotes}
    \end{threeparttable}
    }
\end{table*}
\egroup

The classification accuracies achieved on the \ac{shd}, \ac{ssc}, and \ac{psmnist} test sets, averaged over five independent training runs each, are summarized in \tabref{tab:perf_comparison} together with the throughput, clock frequency, resource, and power figures of the corresponding \acs{asic} and \ac{fpga} implementations. For each dataset, two configurations are evaluated: a throughput-optimized \emph{parallel} variant and a resource-reduced \emph{time-multiplexed} variant. The achieved accuracy of $92.46\pm{}0.69$\% on the \ac{shd} dataset is exceeded by only one prior accelerator, which does so on a substantially larger \ac{fpga} while requiring roughly four times the \acp{lut} and approximately three times the power of our parallel configuration (10.24~W versus 3.325~W). 

To isolate architectural efficiency from differences in technology node and device scale, \tabref{tab:eff_metrics_fpga} and \tabref{tab:eff_metrics_asic} normalize the results of \tabref{tab:perf_comparison} to logic resources, silicon area, and energy. On the \ac{fpga} side, the parallel configuration sustains 112.1~Inf./s/kLUT on the \ac{shd} dataset, a 64\% improvement over the accuracy-wise closest prior work~\cite{nakano_blachut} (68.5~Inf./s/kLUT), while simultaneously offering higher throughput ($\approx{}$6{,}023 versus 5{,}586.59~Inf./s), marginally higher accuracy (92.46\% versus 92.30\%), 34\% fewer \acp{lut}, and no \acs{dsp} blocks. The low-LUT configuration of~\cite{nakano_blachut} achieves a nominally higher 165.8~Inf./s/kLUT, but at 88.78\% accuracy (3.68 percentage points below this work), and is therefore excluded from the comparison (see~\tabref{tab:eff_metrics_fpga}). Among the remaining designs, the proposed parallel configuration delivers the highest throughput density. In energy terms, the configuration reaches $1.81\times{}10^{3}$~Inf./J, ahead of both configurations of~\cite{nakano_blachut}. Notably, all proposed \ac{fpga} implementations operate entirely without \acs{dsp} blocks, preserving these resources for co-processing, and the time-multiplexed \ac{shd} configuration occupies fewer than 21{,}000 \acp{lut}, bringing deployment on mid-range devices within reach.

The Spiker+ implementation~\cite{spiker+} reports the lowest energy per inference among the compared \ac{fpga} designs (232.20~$\mu{}$J/Inf.), however, as noted in \tabref{tab:perf_comparison}, the reported 0.43~W most plausibly accounts for dynamic power only, since it is comparable to the static power draw of the XA7Z020 device alone. Furthermore, this efficiency is achieved at a substantial accuracy penalty (72.99\% versus 92.46\%) and with more than $3\times$ lower throughput (1{,}851.85 versus 6{,}023~Inf./s).

\acs{asic} synthesis was performed using the open-source OpenROAD software~\cite{openroad} and the accompanying \acp{orf}, targeting the openly available ASAP7 \ac{pdk}~\cite{ASAP7}. The complete design flow was executed for the accelerators of all three datasets. To ensure a consistent and reproducible synthesis environment, the container hosted by the Institute for Integrated Circuits and Quantum Computing at the Johannes Kepler University~\cite{iic_osic_tools} was used.\footnote{No tapeout was completed at this point in time. All results are from synthesis and simulation only.} As shown in \tabref{tab:eff_metrics_asic}, the parallel \acs{asic} configuration achieves an energy efficiency of $49.1\times{}10^{3}$~Inf./J on the \ac{shd} dataset ($1.8\times$ that of Loihi~\cite{10558209}), while delivering a $109\times$ higher throughput ($\approx{}$12{,}027 compared to 109.89~Inf./s). Although Loihi's absolute power draw is lower (0.004~W compared to 0.245~W), its energy per inference (36.40~$\mu{}$J/Inf.) exceeds that of the parallel configuration (20.37~$\mu{}$J/Inf.), while the time-multiplexed variant remains competitive at $43.74\mu{}$J/Inf. The higher power draw of the time-multiplexed configuration on the \ac{shd} dataset (0.296~W versus 0.245~W) stems from its $\approx{}1.7\times$ higher operating frequency (1103~MHz versus 642~MHz), whose dynamic power penalty dominates in this case. 

The utilized area of the time-multiplexed configurations remains within 13\% of the parallel variants across all datasets (e.g., 0.0933~mm\textsuperscript{2} versus 0.1073~mm\textsuperscript{2} on the \ac{shd} dataset), since the dynamic reload version requires an additional SRAM for storing the neuron membrane potentials. On the \ac{psmnist} dataset this even results in a marginally larger footprint (0.0885~mm\textsuperscript{2} versus 0.0879~mm\textsuperscript{2}), further highlighting that the time-multiplexed versions are targeted at FPGAs more so than ASICs. The proposed \acs{asic} designs achieve a throughput density of up to $135.8\times{}10^{3}$~Inf./s/mm\textsuperscript{2} on the \ac{psmnist} dataset and clock frequencies of up to 1.1~GHz despite the use of a predictive \ac{pdk}. Relative to their \ac{fpga} counterparts, the \acs{asic} implementations reduce power by more than an order of magnitude (e.g., 0.245~W versus 3.325~W on the \ac{shd} dataset) and energy per inference by 27--37$\times$, demonstrating that a fully open-source design flow can yield competitive \acp{snn} accelerators.

\begin{table}
    \centering
    \begin{threeparttable}
        \caption{Normalized FPGA efficiency metrics.}
        \label{tab:eff_metrics_fpga}
        \begin{tabular}{llrrr}
            \toprule
            Work & Dataset & 
            \shortstack{Inf./s/kLUT} & 
            \shortstack{Inf./J\\($\cdot10^{3}$)} &
            $\mu$J/Inf. \\
            \midrule
            Ours (Par.)                & SHD             & \textbf{112.1}  & \textbf{1.81}  & \textbf{552.05} \\
            Ours (TM)\tnote{*}         & SHD             & 100.4           & 1.25           & 801.07 \\
            \cite{nakano_blachut}      & SHD (high acc.) & 68.5            & 1.42           & 705.26 \\
            \cite{nakano_blachut}      & SHD (low-LUT)   & $^\dagger$165.8 & 1.51           & 663.25 \\
            \cite{spiker+}             & SHD             & N/R             & $^\dagger$4.31 & $^\dagger$232.20 \\ 
            \midrule
            Ours (Par.)        & SSC  & \textbf{30.5} & \textbf{0.44} & \textbf{2274.21} \\
            Ours (TM)\tnote{*} & SSC  & 27.1          & 0.34          & 2961.49          \\
            \midrule
            Ours (Par.)        & PSMNIST & 107.8          & \textbf{1.38} & \textbf{725.12}\\
            Ours (TM)\tnote{*} & PSMNIST & \textbf{136.4} & 1.17          & 852.85         \\
            \bottomrule
        \end{tabular}
        \begin{tablenotes}
            \item[]$*$ Time-multiplexed (TM) using dynamic neuron reuse.
            \item[]N/R Not reported in original work. 
            \item[]$^\dagger$ Result disregarded because of the significantly lower accuracy of the referred work.
        \end{tablenotes}
    \end{threeparttable}
\end{table}

\begin{table}
    \centering
    \begin{threeparttable}
        
    \caption{Normalized ASIC efficiency metrics.}
    \label{tab:eff_metrics_asic}
        \begin{tabular}{llrrr}
            \toprule
            Work & Dataset & \shortstack{Inf./s/mm\textsuperscript{2}\\($\cdot10^{3}$)} & 
            \shortstack{Inf./J\\($\cdot10^{3}$)} & \shortstack{$\mu$J/Inf.\\ ($\cdot10^{3}$)} \\
            \midrule
            Ours (Par.)        & SHD  & \textbf{112.1} & \textbf{49.1} & \textbf{20.37}\\
            Ours (TM)\tnote{*} & SHD  & 72.5           & 22.9          & 43.74 \\
            \cite{10558209}    & SHD  & N/R            & 27.5          & 36.40 \\
            \midrule
            Ours (Par.)        & SSC  & \textbf{34.6}     & \textbf{13.9} & \textbf{71.96} \\
            Ours (TM)\tnote{*} & SSC  & 15.4              & 6.2           & 160.25         \\
            \midrule
            Ours (Par.)        & PSMNIST & \textbf{135.8} & \textbf{51.2} & \textbf{19.52} \\
            Ours (TM)\tnote{*} & PSMNIST & 70.3           & 28.0          & 35.67          \\
            \bottomrule
        \end{tabular}
        \begin{tablenotes}
            \item[]$*$ Time-multiplexed (TM) using dynamic neuron reuse.
            \item[]N/R Not reported in original work. 
        \end{tablenotes}
    \end{threeparttable}
\end{table}

\section{Conclusion}
\label{sec:conclusion}
This work evaluates three spike train representation methodologies for digital neuromorphic hardware and demonstrates that \ac{dte} significantly reduces the total bit width required to represent spike trains compared to absolute time representation or \mbox{1-bit} equidistantly sampled spike trains. We further introduce a mathematical proof for merging separate \ac{dte} spike trains and propose a corresponding low-complexity hardware implementation.

Building on this representation, we present a set of modular hardware components that flexibly compose into custom \ac{snn} accelerators, supporting recurrent and feedforward networks with learned axonal delays and thus current state-of-the-art \acp{snn}, in both throughput-optimized and resource-reduced configurations. A custom Python framework generates the top-level VHDL from JSON configuration files, streamlining the derivation of accelerators for new datasets. Notably, the resulting designs are entirely multiplication-free, requiring neither multipliers nor \ac{dsp} blocks.

To support reproducibility and further research, the complete VHDL source code is available in a public GitHub repository. Synthesis results for the Xilinx ZCU102 \ac{fpga} demonstrate up to $\approx{}6{,}023$~Inf./s at a throughput density of 112.1~Inf./s/kLUT on the \ac{shd} dataset, surpassing prior accelerators of comparable accuracy. \ac{asic} implementations synthesized with the open-source OpenROAD flow and the predictive ASAP7 \ac{pdk} reach up to $\approx{}12{,}027$~Inf./s with clock frequencies exceeding 1.1~GHz and surpass the energy efficiency of Intel's Loihi~\cite{10558209} by $1.8\times$ ($49.1\times{}10^{3}$ versus $27.5\times{}10^{3}$~Inf./J). 


\ifappendix
\appendix 

\subsection{Problem Statement}
The objective is to merge $N$ independent sequences of time durations $\mathcal{S}_m = \{ \Delta S_{m,i} \}_{i=1}^{\infty}$ into a single sequence $\Delta Y_k$, representing the inter-event intervals of the ordered union of all source processes.

%

\subsection{Literature Analysis}
The algorithm formalizes the recursive implementation of the superposition of point processes. 

\begin{itemize}
    \item \textbf{Stochastic Modeling:} The foundation lies in Poisson process superposition. Kingman \cite{kingman-poisson-processes} established that the merger of independent arrival streams forms a point process where inter-arrival times are governed by the minimum of current residuals.
    \item \textbf{Digital Forensics:} When absolute clocks are unreliable, forensics relies on unifying relative time frames. Stevens \cite{stevens2004} and Dreier et al. \cite{beyond-timestamps} emphasize integrating implicit timing to reconstruct coherent timelines without a common clock.
\end{itemize}

\textbf{Theorem.} The sequence of merged inter-event durations $\Delta Y_k$ can be uniquely generated by an inductive recursion over a residual vector $\mathbf{R}_k = (R_{1,k}, \dots, R_{n,k})$ representing the time until the next event in each active stream.

\begin{theorem}[Correctness of N-Stream Differential Merge] 
	Let $\mathcal{S}_m = \{ \Delta S_{m,i} \}_{i\ge 1}$ be $N$ sequences of nonnegative inter-event durations (one per stream $m=1,\dots,N$). Define absolute event times per stream by
	\[
	a_{m,1} := \Delta S_{m,1},
	\quad
	a_{m,i+1} := a_{m,i} + \Delta S_{m,i+1} \quad \text{for } i\geq 1,
	\]
	and, if stream $m$ is finite, set $a_{m,i} := +\infty$ for all $i$ beyond its last event.
	
	Let $\{ y_k \}_{k\ge 1}$ be the nondecreasing sequence obtained by sorting the \emph{multiset} $\biguplus_{m=1}^N \{ a_{m,i} \}_{i\ge 1}$ with multiplicity, and set $y_0 := 0$. Consider the residual-time vector $\mathbf{R}_k = (R_{1,k},\dots,R_{N,k})$ and the differential output $\Delta Y_k$ generated as follows:
	\begin{align}
		R_{m,1} &:= \Delta S_{m,1} = a_{m,1} - y_0, \qquad m=1,\dots,N, \label{eq:init} \\
		\Delta Y_k &:= \min_{1\le m\le N} R_{m,k}, \qquad k\ge 1, \label{eq:out} \\
		\mathcal{F}_k &:= \bigl\{ m : R_{m,k} = \Delta Y_k \bigr\} \quad \text{(firing set)}, \label{eq:firing} \\
		R_{j,k+1} &:= R_{j,k} - \Delta Y_k, \qquad j\notin \mathcal{F}_k, \label{eq:update-nonfiring} \\
		R_{m,k+1} &:= \Delta S_{m,\mathrm{next}(m,k)}, \qquad m\in \mathcal{F}_k, \label{eq:update-firing}
	\end{align}
	where $\mathrm{next}(m,k)$ denotes the (per-stream) index of the event that fired for stream $m$ at step $k$ (or is set so that $R_{m,k+1}=+\infty$ if the stream is exhausted).
	
	Then, for all $k\ge 1$, $\Delta Y_k := y_k - y_{k-1}$.
	
	In particular, $\Delta Y_k = 0$ if and only if $y_k = y_{k-1}$ (simultaneous events across streams or within a stream).
\end{theorem}

\begin{proof}
	We prove by induction on $k\ge 1$ that:
	\begin{equation}
		\label{eq:residual-invariant}
		R_{m,k} := a_{m,\mathrm{next}(m,k)} - y_{k-1}
		\quad\text{for all } m=1,\dots,N,
	\end{equation}
	and that the output rule \eqref{eq:out} produces $\Delta Y_k = y_k - y_{k-1}$.
	
	We first clarify $\mathrm{next}(m,k)$. For each stream $m$ and step $k\ge 1$, let $\mathrm{next}(m,k)$ be the minimum index $i$ such that $a_{m,i} \ge y_{k-1}$, with the convention that, if $a_{m,i}=y_{k-1}$ and stream $m$ fired at step $k-1$, then $\mathrm{next}(m,k)=i+1$ (or $+\infty$ if exhausted). Operationally, the update rule \eqref{eq:update-firing} enforces exactly this advancement.
	
	Base case ($k=1$). By \eqref{eq:init}, $R_{m,1}=a_{m,1}-y_0$, so \eqref{eq:residual-invariant} holds for $k=1$ with $\mathrm{next}(m,1)=1$. The output rule \eqref{eq:out} gives
	\begin{align*}
	\Delta Y_1 &:= \min_m R_{m,1} \\ 
    &= \min_m (a_{m,1}-y_0)\\
	&= \bigl(\min_m a_{m,1}\bigr) - y_0 \\
    &= y_1 - y_0.
	\end{align*}
	
	Thus the claim holds for $k=1$.
	
	Inductive step. Fix $k\ge 1$ and assume \eqref{eq:residual-invariant} holds at step $k$, i.e.,
	\[
	R_{m,k} := a_{m,\mathrm{next}(m,k)} - y_{k-1}
	\quad \text{for all } m.
	\]
	By the output rule \eqref{eq:out} and the invariant,
	\begin{align*}
		\Delta Y_k
		&= \min_m R_{m,k} \\
		&= \min_m \bigl( a_{m,\mathrm{next}(m,k)} - y_{k-1} \bigr) \\
		&= \bigl( \min_m a_{m,\mathrm{next}(m,k)} \bigr) - y_{k-1} \\
		&= y_k - y_{k-1},
	\end{align*}
	where $y_k$ is, by definition, the next element in the nondecreasing merge of the multiset $\biguplus_m \{a_{m,i}\}$.
	
	It remains to show that \eqref{eq:residual-invariant} holds at step $k+1$.
	
	Non-firing streams. For any $j\notin \mathcal{F}_k$, by \eqref{eq:update-nonfiring} and the inductive hypothesis, 
	\begin{align*}
	R_{j,k+1} &= R_{j,k} - \Delta Y_k \\
    &= \bigl( a_{j,\mathrm{next}(j,k)} - y_{k-1} \bigr) - (y_k - y_{k-1}) \\
    &= a_{j,\mathrm{next}(j,k)} - y_k. 
	\end{align*}
	Since $j$ did not fire at step $k$, its next event index remains the same, $\mathrm{next}(j,k+1)=\mathrm{next}(j,k)$, and hence 
	\[ 
	R_{j,k+1} = a_{j,\mathrm{next}(j,k+1)} - y_k, 
	\] 
	which is \eqref{eq:residual-invariant} at step $k+1$ for stream $j$.
	
	Firing streams. For any $m\in \mathcal{F}_k$, we have $R_{m,k}=\Delta Y_k$, so 
	\begin{align*}
	a_{m,\mathrm{next}(m,k)} - y_{k-1} &= \Delta Y_k \\
    &= y_k - y_{k-1} \quad \Longrightarrow \quad a_{m,\mathrm{next}(m,k)} \\
    &= y_k. 
	\end{align*}
	Thus the event indexed by $\mathrm{next}(m,k)$ occurs exactly at $y_k$. Advancing the per-stream index to the next event (or to $+\infty$ if exhausted), we have 
	\[ 
	\mathrm{next}(m,k+1) = \mathrm{next}(m,k) + 1, 
	\] 
	and the update \eqref{eq:update-firing} gives 
	\begin{align*}
	R_{m,k+1} &= \Delta S_{m,\mathrm{next}(m,k)} \\
    &= a_{m,\mathrm{next}(m,k)+1} - a_{m,\mathrm{next}(m,k)} \\ 
    &= a_{m,\mathrm{next}(m,k+1)} - y_k, 
    \end{align*}
	which is \eqref{eq:residual-invariant} at step $k+1$ for stream $m$.
	
	Together, these cases show that \eqref{eq:residual-invariant} is preserved from $k$ to $k+1$, completing the induction.
	
	Simultaneous events. If multiple streams (or multiple events within a stream) have $a_{m,i}=y_k$, then all such streams lie in $\mathcal{F}_k$. The updates in \eqref{eq:update-firing} advance each of them past time $y_k$. If further events still occur at $y_k$, the next step yields $\Delta Y_{k+1}=0$, and so on, producing as many zero outputs as the multiplicity at $y_k$. Hence the algorithm emits the nondecreasing merged sequence of absolute times via the differences $\Delta Y_k$.
	
	Exhausted streams. If a stream is finite, setting $a_{m,i}=+\infty$ beyond its last event implies $R_{m,k}=+\infty$ once exhausted, which never wins the minimum. Equivalently, an implementation may remove exhausted streams from the priority queue; both views are consistent with the algebra above.
	
	Therefore, for all $k\ge 1$, $\Delta Y_k = y_k - y_{k-1}$, as claimed.
\end{proof}

\fi

\section*{Acknowledgments}
This research received funding from the Austrian Research Promotion Agency (FFG) and the Austrian ministry for Climate Action, Environment, Energy, Mobility, Innovation and Technology (BMK) under project no. FO999899263 and funding from the Chips Joint Undertaking under project no. 101112274 (ISOLDE) and its members Austria, Czechia, France, Germany, Italy, Romania, Spain, Sweden, Switzerland. 

This work is furthermore supported by: the COMET-K2 ``Center for Symbiotic Mechatronics'' of the Linz Center of Mechatronics (LCM), funded by the Austrian federal government and the federal state of Upper Austria. 

This work has been supported by Silicon Austria Labs (SAL), owned by the Republic of Austria,  the Styrian Business Promotion Agency (SFG), the federal state of Carinthia, the Upper Austrian Research (UAR), and the Austrian Asso­ci­a­tion for the Elec­tric and Elec­tronics Industry (FEEI).

\bibliographystyle{IEEEtran}
\balance
\bibliography{bibliography.bib}

\vfill

\end{document}